%% file: root.tex
\documentclass[letterpaper, 10 pt, conference]{ieeeconf}  % Comment this line out if you need a4paper

\usepackage{graphicx}
\usepackage{amsfonts}
\usepackage[T1]{fontenc}
\usepackage{amsmath}
\usepackage{amssymb}
\usepackage{multirow}
\usepackage[table]{xcolor}
\usepackage{cite}
\usepackage{booktabs}
\usepackage{bm}
\usepackage[font=small]{caption}
\usepackage{lipsum}
\usepackage{pifont}
\usepackage{makecell}
\usepackage{titlesec}
\usepackage{xspace}
\usepackage{hyperref}

\IEEEoverridecommandlockouts                              % This command is only needed if 
\newcommand{\algname}{A-SLIP\xspace}
\title{\LARGE \bf
A-SLIP: Acoustic Sensing for Continuous In-hand Slip Estimation
}

\author{
Uksang~Yoo\textsuperscript{*,1},
Yuemin~Mao\textsuperscript{*,1},
Jean~Oh\textsuperscript{1},
Jeffrey~Ichnowski\textsuperscript{1}
\thanks{
\textsuperscript{1}Robotics Institute, Carnegie Mellon University, USA;
\textsuperscript{*}Equal contribution.
}
}

\begin{document}

\maketitle
\thispagestyle{empty}
\pagestyle{empty}

%We present \algname, an embedded acoustic sensing system for continuous planar slip vector estimation using piezoelectric microphones integrated into a parallel-jaw gripper. Each sensor has a textured silicone contact pad to promote structured contact-induced vibrations. \algname processes synchronized multi-channel audio as log-mel spectrograms with a lightweight convolutional network. %with learned channel attention and temporal attention pooling. [I think we can leave this detail out to save some space -JI]
%%%%%%%%%%%%%%%%%%%%%%%%%%%%%%%%%%%%%%%%%%%%%%%%%%%%%%%%%%%%%%%%%%%%%%%%%%%%%%%%
\begin{abstract}
Reliable in-hand manipulation requires accurate real-time estimation of the slip between a gripper and a grasped object. Existing tactile sensing approaches based on vision, capacitance, or force-torque measurements face fundamental trade-offs in form factor, durability, and their ability to jointly estimate slip direction and magnitude. We present \algname, a multi-channel acoustic sensing system integrated into a parallel-jaw gripper for estimating continuous slip in the grasp plane. The \algname sensor consists of piezoelectric microphones positioned behind a textured silicone contact pad to capture structured contact-induced vibrations. The \algname model processes synchronized multi-channel audio as log-mel spectrograms using a lightweight convolutional network, which jointly predicts the presence, direction, and magnitude of the slip. Across experiments with robot- and externally-induced slip conditions, the finetuned four-microphone configuration achieves a mean absolute directional error of 14.1 degrees, outperforms baselines by up to 12\% in detection accuracy, and reduces directional error by 32\%. Compared with single-microphone configurations, the multi-channel design reduces directional error by 64\% and magnitude error by 68\%, underscoring the importance of spatial acoustic sensing in resolving slip direction ambiguity. We further evaluate \algname in closed-loop reactive control, and find that it enables reliable and low-cost real-time estimation of in-hand slip. Project videos and additional details are available at  \href{https://a-slip.github.io/#}{a-slip.github.io}.
\end{abstract}

%\\algname{} improves slip detection accuracy by approximately 12\% relative to a classical SVM baseline and reduces directional estimation error by about 26\%. Compared to single-microphone models, the multi-channel design reduces directional error by up to 32\%, highlighting the importance of spatial acoustic sensing for resolving slip direction. We further demonstrate that \algname{} enables real-time closed-loop responses to in-hand slip, supporting reactive grasp stabilization.

%%%%%%%%%%%%%%%%%%%%%%%%%%%%%%%%%%%%%%%%%%%%%%%%%%%%%%%%%%%%%%%%%%%%%%%%%%%%%%%%
% \section{Introduction}
% Reliable in-hand manipulation requires a robot to maintain stable and controlled contact with grasped objects throughout a task. A challenge in achieving this is the real-time detection and estimation of slip, defined as the relative motion between gripper fingers and object surface. If unobserved and responded to, the slip may result in dropped objects, or unintended disturbances to the environment. Slip is particularly difficult to handle because it is transient, directionally variable, and frequently occluded by the robot gripper, while also occurring faster than a reactive control loop can compensate without advance sensing. Addressing this challenge is therefore crucial for robust robot manipulator deployment in unstructured real-world settings.

\section{Introduction}
Reliable in-hand manipulation requires a robot to maintain stable and controlled contact with grasped objects throughout a task. A central challenge is real-time detection and estimation of slip, defined as relative motion between the gripper fingers and the surface of the object. When a slip goes unobserved or uncorrected, it can lead to dropping objects, task failures, or unintended disturbances to the environment. Detecting slip is particularly challenging because it is transient, directionally varying, and often occluded by the gripper. Moreover, slip events frequently occur on timescales faster than a purely reactive control loop can compensate without advanced sensing. As a result, reliable slip estimation is a critical capability for robust in-hand manipulation and the deployment of robot manipulators in unstructured real-world environments.

\input{floats/fig_front}

Researchers have approached in-hand slip sensing through various modalities. Wrist-mounted force-torque sensors can detect the onset of slip through observing changes in the measured wrench, but they provide ambiguous signals for slip direction and are sensitive to external contact disturbances unrelated to the slip. Capacitive and resistive tactile sensor arrays offer spatially resolved pressure measurements, and prior work has demonstrated their ability to infer slip from shear and pressure redistribution patterns~\cite{yuan2015measurement}. However, these sensors are sensitive to wear, require complex fabrication, and degrade over repeated use due to difficult-to-model soft sensor phenomena such as creep and hysteresis. Vision-based tactile sensors such as GelSight~\cite{yuan2017gelsight} and DIGIT~\cite{lambeta2020digit} embed cameras beneath a deformable gel surface and can capture rich contact geometry with high spatial resolution. However, vision-based tactile sensors commonly suffer from bulky form factors, low data-acquisition rates, limited scalability and sensor coverage, and low durability under repeated contact due to a thin spectral coating, restricting their practical deployment.  

A promising alternative is acoustic sensing. When a slip occurs at the gripper-object interface, friction and surface asperities generate structure-borne vibrations that propagate through the sensor body and can be captured by microphones embedded in the fingers. Prior work has shown that acoustic signals carry information about contact events and surface properties during manipulation~\cite{mao2025visuo, lee2025sonicboom, zoller2020active}, and that piezoelectric microphones can detect the onset of slip~\cite{mao2025hearingslide} in the presence of robot operating noise. However, existing acoustic approaches have largely been limited to binary slip detection and do not address the estimation of slip direction or magnitude, which are necessary for closed-loop grasp correction. Moreover, prior work has generally not explored multi-channel acoustic fusion or learning-based approaches for continuous slip vector estimation.

In this work, we present Acoustic Sensing for Learning In-hand slip Parameters (\algname), an embedded acoustic sensing system for slip direction and magnitude estimation. We design a low-profile gripper sensor consisting of a textured silicone contact surface with embedded piezoelectric microphones. The textured silicone promotes structured vibrations at the contact interface during slip, while the piezoelectric microphones provide broadband sensitivity to structure-borne sound with minimal footprint. Compared to vision-based tactile sensors, the \algname{} design requires no optics, illumination, or cameras, resulting in a sensor that is more compact, durable, and low cost. Building on this hardware, \algname learns to map synchronized multi-microphone spectrograms to a continuous planar slip vector that jointly encodes slip event, direction, and magnitude within a unified prediction framework.

The contributions of this work are as follows:
\begin{itemize}[leftmargin=*] %,labelwidth=12pt,itemindent=12pt]
    \item \algname sensor, a low-profile and low-cost acoustic gripper sensor system based on a textured silicone contact surface with embedded piezoelectric microphones that is durable and suitable for real-world deployment across parallel-jaw gripper platforms.
    
    \item \algname model, a slip prediction network that %fuses synchronized multi-channel spectrograms using learned channel attention and temporal attention pooling, and 
    jointly estimates slip presence, magnitude, and direction through a unified multi-objective formulation with a two-stage pre-training and finetuning strategy. % consisting of pretraining and motion-capture-supervised finetuning.
    
    \item Experiments with \algname{} on real-time estimation of in-hand slip and  reactive closed-loop control. %experiments, where predicted slip vectors are used to detect slip events and guide corrective motion across multiple objects and externally induced disturbances.
\end{itemize}

\section{Related Work}

Prior work explored three relevant directions: tactile sensing hardware, %for manipulation,
slip estimation methods, % and their sensing requirements, 
and acoustic sensing as an emerging modality for contact-rich robotics.

\subsection{Tactile Sensing}
Touch is a fundamental modality underlying human dexterity, and tactile sensing has been extensively studied to endow robots with similar capabilities. Force-torque sensors characterize touch by measuring contact forces and moments directly~\cite{cao2021six, shademan2016supervised, nadeau2022fast, suresh2021tactile}; vision-based tactile sensors infer touch from high-resolution surface deformation~\cite{ she2021cable, alspach2019soft, lambeta2020digit, oller2023manipulation}; capacitive and magnetic-based sensors localize contact by measuring changes in electric or magnetic fields induced by deformation or proximity~\cite{hellebrekers2020soft, wi2026tactalign, liu2024material, yao2020highly}. Multimodal tactile sensors  combine sensing modalities to extract more comprehensive information from physical interactions~\cite{mao2024multimodal, higuera2025tactile}. Prior works have demonstrated that tactile sensing improves manipulation performance in various domains, such as object geometry recovery~\cite{huang2025gelslam}, object material property estimation~\cite{han2025estimating}, and dexterous in-hand manipulation~\cite{wang2025lessons}.

However, existing sensors often remain difficult to deploy in general-purpose manipulation due to form factor, calibration complexity, and cost. In contrast, we propose acoustic sensing, which characterizes contact through structure-borne mechanical vibrations generated by physical interaction. This tactile sensing modality is compact, low-cost, and capable of delivering high-frequency, low-latency signals suitable for real-time control.

\subsection{Slip Estimation}
Slip estimation has been approached through sensing modalities with distinct trade-offs in signal richness, latency, and deployability. Wrist-mounted force-torque sensors can detect slip onset via abrupt changes in the measured wrench~\cite{romeo2020slipdetectionsurvey, stachowsky2016slip}, but provide limited directional information and often conflate slip-induced loads with external contact disturbances~\cite{yuan2015measurement}. Tactile sensor arrays enable spatially resolved estimation by tracking shear and pressure redistribution across the contact patch~\cite{huang20253d}, while vision-based tactile sensors such as GelSight~\cite{yuan2017gelsight} and DIGIT~\cite{lambeta2020digit} further extend this capability to high-resolution contact geometry reconstruction. Learning-based methods, including convolutional and recurrent architectures for slip classification from tactile sequences~\cite{yuan2017connecting} and self-supervised approaches that reduce labeling requirements~\cite{higuera2025sparsh}, have improved prediction from raw tactile streams. However, these approaches often rely on vision-based tactile sensors with thin compliant gel surfaces and optical coatings that are susceptible to wear, hysteresis, and performance degradation under repeated shear~\cite{lambeta2020digit}. As a result, many datasets emphasize contact events or controlled interactions rather than sustained slip.

%Learning-based methods, including convolutional and recurrent architectures for slip classification from tactile sequences~\cite{yuan2017connecting} and self-supervised strategies to reduce labeling requirements~\cite{higuera2025sparsh}, have improved prediction from raw sensor streams,  %but remain tightly coupled to specific hardware and often struggle to generalize across gripper platforms, contact geometries, or surface properties.

Acoustic sensing offers a compelling alternative for slip estimation. Piezoelectric microphones are rigid, wear-resistant, and can fit into smaller form factors, while slip-induced vibrations propagate through the gripper on timescales faster than vision-based feedback can resolve~\cite{rangwala1988application}. Prior acoustic approaches have largely focused on binary slip detection~\cite{chen2018tactile} or contact event recognition~\cite{lu2023active} and do not estimate slip direction or magnitude. \algname addresses these limitations through a multi-channel acoustic sensing pipeline and a learning-based architecture for continuous planar slip vector estimation, enabling actionable feedback for closed-loop manipulation.

\subsection{Acoustic Sensing for Manipulation}
Acoustic sensing in robotics can be categorized along two orthogonal dimensions. The first is the \emph{propagation medium}: airborne acoustics captures sounds transmitted through air, typically corresponding to human-audible interaction cues such as pouring~\cite{liang2019pouring}, whereas structure-borne acoustics captures mechanical vibrations transmitted through rigid bodies and often encodes contact phenomena that are imperceptible without instrumentation. The second is the sensing mode: \emph{passive sensing} listens to naturally occurring signals during interaction~\cite{lee2025sonicboom, andrussow2025adding}, while \emph{active sensing} emits a probing signal and analyzes the response~\cite{yoo2024poe, zhang2025vibecheck}. Prior work has leveraged acoustic sensing in manipulation both to learn task-relevant representations, such as material properties or contact states, for downstream performance~\cite{lu2023active}, and as an auxiliary modality within end-to-end learning pipelines~\cite{liu2024maniwav}.

\input{floats/fig_design}

% Slip events are brief, contact-localized, and often weakly observable through airborne sound, but they generate rich, high-frequency structure-borne vibrations that encode both the presence and direction of relative motion at the contact interface. Recovering this information requires sensing configurations and learning architectures that go beyond binary event detection or global audio cues.

% \algname leverages passive structure-borne acoustic sensing %using piezoelectric microphones embedded in the gripper, 
% for robust capture of slip-induced vibrations even in the presence of robot actuation noise. %By leveraging synchronized multi-channel measurements and learning-based fusion,
% \algname extracts directional asymmetries in vibrations to estimate a continuous planar slip vector. %This establishes acoustic sensing not merely as a complementary perception signal, but as a precise and actionable feedback modality for real-time, closed-loop in-hand manipulation.

Slip events generate brief, contact-localized, high-frequency structure-borne vibrations that encode both the presence and direction of relative motion at the contact interface. \algname leverages passive structure-borne acoustic sensing with piezoelectric microphones embedded in the gripper and learns directional asymmetries in slip-induced vibrations to estimate a continuous planar slip vector encoding both slip direction and magnitude.

% \subsection{Acoustic Sensing for Manipulation}
% Acoustic sensing in robotics can be broadly categorized along two dimensions. The first dimension distinguishes between airborne acoustics, which captures sounds propagating through air and typically focuses on human-audible interaction cues (e.g. pouring) [], and structure-borne acoustics, which captures mechanical vibrations transmitted through rigid bodies and often contains information imperceptible to humans without instrumentation []. The second dimension distinguishes between passive acoustic sensing, where the system only listens to naturally occurring signals during interaction [], and active sensing, where the system emits a probing signal and analyzes the response []. In application to robotic manipulation, acoustic sensing has mainly been used to learn task-relevant representations, such as material properties and contact states, or to improve downstream task performance[], or as an additional modality integrated into end-to-end learning pipelines [].

% We employ structure-borne acoustic since slip events produce minimal airborne sound but generate informative high-frequency vibrations detectable through the structure, and passive sensing as prior work has demonstrated that such vibrations remain measurable in the presence of robot noise []. 

\section{Problem Formulation}

The problem is to infer in-hand planar object slip presence, direction, and magnitude from acoustic observations during grasped manipulation. %Consider a parallel-jaw gripper in persistent contact with an object. Slip, when it occurs, is constrained to the local contact plane of the gripper fingers and can be modeled as a latent planar motion signal.
Let $\mathbf{X}_t^n=\{\mathbf{x}_t^1, \mathbf{x}_t^2, ..., \mathbf{x}_t^n\}$ denote synchronized audio measurements of $n$ microphones embedded in the gripper at time $t$, where $\mathbf{x}_t^i \in \mathbb{R}^{T \times F}$ represents a short-time spectral representation over a temporal window of duration $T$. These measurements provide indirect observations of vibrations at the gripper-object interface.

We define the slip state at time $t$ as a planar slip vector
\[
\mathbf{v}_t = (v_x, v_z) \in \mathbb{R}^2,
\]
where the direction of $\mathbf{v}_t$ encodes the instantaneous direction of slip in the gripper plane and its magnitude $\lVert\mathbf{v}_t\rVert$ corresponds to the intensity of slip. The no-slip condition is captured naturally by $\mathbf{v}_t = \mathbf{0}$.

The goal is to learn a function
% \[
% f_\theta : \left(\mathbf{X}_t^1, \mathbf{x}_t^2\right) \mapsto \mathbf{v}_t
% \]
\[
f_\theta : \mathbf{X}_t^n \mapsto \mathbf{v}_t,
\]
parameterized by $\theta$. %that estimates the instantaneous slip vector from multi-channel acoustic observations. % [Since X and v are already defined, you don't need to define them again. -JI]
As $\mathbf{X}_t^n$ depends on the sensor design and microphone placement, we also seek a sensor configuration to minimize slip-estimation error.

% This formulation jointly captures slip detection (via $\|\mathbf{v}_t\| > 0$) and slip direction estimation within a unified continuous prediction framework.

\section{Methods}

\algname achieves state-of-the-art acoustic in-hand slip estimation through a unique recipe of hardware design, network architecture, and dataset curation. 

% \begin{figure}[t!]
%     \centering
%     \includegraphics[width=\linewidth]{figures/texture.pdf}
%     \caption{%
%     \textbf{Pad texture comparison.} Angle prediction MAE (degrees) over training steps for smooth and textured contact pads over the validation set. The textured pad converges to significantly lower directional error than the smooth pad, suggesting that surface asperities generate more directionally informative structure-borne vibrations during slip compared to smooth textures.
%     }
%     \label{fig:texture}
% \end{figure}

% This problem is challenging due to the highly indirect relationship between acoustic signals and slip dynamics, the presence of sensor- and contact-dependent noise, and the ambiguity introduced by symmetric contact geometries. Nevertheless, multi-channel acoustic sensing provides complementary spatial cues that can be exploited to resolve slip direction and magnitude in real time.

\subsection{Hardware Design}
We propose a low-profile acoustic gripper sensor design composed of two primary components: a molded silicone contact pad and a rigid finger-mounted holder with embedded piezoelectric microphones (Fig.~\ref{fig:design}).

% \subsubsection{Silicone Contact Pad.} 
% To promote conformal contact with grasped objects while remaining sufficiently stiff to efficiently transmit structure-borne vibrations to the embedded microphone, we chose a two-part platinum-cure liquid silicone rubber (Shore 30A) for the contact pad. We mixed the silicone at a 1:1 volume ratio and cast it with a custom 3D-printed mold that imprints a regular surface texture onto the contact face, inspired by prior work on vibration-inducing surface texture designs for other tactile modalities~\cite{dai2022design}. This texture introduces controlled surface asperities that modulate contact-induced vibrations during slip, yielding richer and more directionally informative acoustic signatures than a smooth contact surface. After curing, we demolded the pad and bonded it to the rigid holder with silicone adhesive to ensure strong mechanical and acoustic coupling between the contact surface and sensor body.

To promote conformal contact with grasped objects while remaining sufficiently stiff to efficiently transmit structure-borne vibrations to the embedded microphone, we fabricate the contact pad with a two-part platinum-cure liquid silicone rubber (Shore 30A). We mix the silicone at a 1:1 volume ratio and cast it with a custom 3D-printed mold. After curing, we demold the pad and bond it to the rigid holder with silicone adhesive to ensure strong mechanical and acoustic coupling between the contact surface and sensor body.

Inspired by prior work on vibration-inducing surface textures for other tactile modalities~\cite{dai2022design}, we design two mold variants to produce either a contact face with imprinted regular textures or a smooth contact surface (Fig.~\ref{fig:design}, center). In experiments the textured pad shows a 62.90\,\% reduction in directional MAE compared to the smooth pad, suggesting that the textured surface introduces controlled asperities that modulate contact-induced vibrations and produce richer, more directionally informative acoustic signatures.

%\textbf{Microphone Mounting and Configurations.} 
% Each rigid holder houses one or more piezoelectric microphones mounted directly behind the silicone pad to maximize sensitivity to contact-induced structure-borne vibrations. 
Each rigid finger-mounted holder houses one or more microphones positioned flush with the holder’s top surface. After bonding the silicone pad with the holder, the microphones sit directly beneath the pad, maximizing sensitivity to contact-induced structure-borne vibrations. In experiments (Sec.~\ref{sec:system_ablation}), we evaluate four microphone configurations (Fig.~\ref{fig:design}, right) corresponding to different holder designs.

% We evaluate four microphone configurations (Fig.~\ref{fig:design}). The \emph{2-microphone (centered)} configuration has one microphone centered behind each finger pad. The \emph{2-microphone (corners)} configuration has microphones at opposing corners across the two fingers. The \emph{2-microphone (same-finger)} configuration has both microphones embedded in a single finger. The \emph{4-microphone} configuration has two spatially distributed microphones per finger to increase contact-region coverage and capture directional vibration asymmetries.

%we evaluate two hardware configurations: a \textit{2-microphone configuration}, with one microphone embedded in each finger (two total), and a \textit{4-microphone configuration}, in which two microphones are embedded per finger in a spatially distributed arrangement. The 4-microphone configuration provides increased spatial coverage of the contact region and enables the capture of vibration asymmetries that encode slip direction. In both configurations, microphones are flush-mounted against the rear surface of the silicone pad and rigidly secured within the holder to minimize relative motion artifacts.

% \subsubsection{}{Integration with the Gripper.} 
We mount the assembled sensor on both fingers of a parallel-jaw gripper, replacing the default contact surfaces. The rigid holders match the gripper finger mounting geometry, allowing drop-in installation without structural modification. Microphone signals route through thin-gauge wires along the finger body to an external data acquisition mixer that synchronizes multi-channel audio captured at a fixed sampling rate. The resulting sensor adds minimal bulk to the gripper profile and preserves workspace clearance. %, and enables deployment across different parallel-jaw gripper platforms.

\input{floats/fig_model}

\subsection{Slip Prediction Model}
Slip events manifest as brief broadband friction-induced vibrations at the contact interface. To capture both the frequency content and temporal evolution of these vibrations, we represent each microphone signal as a log-mel spectrogram computed over 200\,ms windows. Each input sample is a tensor $\mathbf{X} \in \mathbb{R}^{n \times M \times T}$, where the $n$ channels correspond to the $n$ microphones, $M$ is the number of mel-frequency bins, and $T$ is the number of time frames. Spectrograms are normalized using dataset-level mean and variance statistics to reduce sensitivity to gain and contact variability.

\algname's slip prediction network is a convolutional architecture designed to preserve fine-grained temporal cues critical for slip direction estimation. A learnable channel attention mechanism first fuses the multi-microphone streams: frequency-averaged spectrograms are passed through a lightweight temporal convolutional gating network that predicts per-channel weights, allowing the model to emphasize microphone with stronger slip-related cues. The fused representation is then processed by a stack of 2D convolutional layers interleaved with batch normalization, ReLU activations, dropout, and frequency-only max pooling to preserve temporal resolution. Subsequent 1D temporal convolution layers capture short-term dynamics associated with slip onset and direction changes. Finally, a learned temporal attention pooling mechanism aggregates features into a fixed-length latent vector, which is passed to three prediction heads: a slip classification head outputting $p(\text{slip})$, a magnitude regression head, and a direction head predicting a unit-normalized 2D vector as shown in Fig.~\ref{fig:architecture}.

\input{floats/fig_system}

We train the network with a multi-objective loss that jointly supervises slip detection, magnitude estimation, and direction estimation. Let $\mathbf{v}^* = (v_x^*, v_z^*) \in \mathbb{R}^2$ denote the ground-truth slip vector defined in the slip plane of the parallel jaw gripper and $\hat{\mathbf{v}} \in \mathbb{R}^2$ the predicted vector. We supervise slip presence with a binary cross-entropy loss
\[
\mathcal{L}_{\text{slip}} = \text{BCE}\left(p(\text{slip}),\, \mathbf{1}\left[\|\mathbf{v}^*\| > \epsilon\right]\right),
\]
where $\epsilon$ defines the slip magnitude threshold to label a sample as in slip.

We supervise slip magnitude with a Huber loss applied only on frames labeled as slip,
\[
\mathcal{L}_{\text{mag}} = \text{Huber}\left(\|\hat{\mathbf{v}}\|,\, \|\mathbf{v}^*\|\right).
\]
%Slip magnitude is standardized using training-set statistics and optimized with a Huber loss,

% Slip direction is supervised using a cosine similarity loss,
% \[
% \mathcal{L}_{\text{dir}} = 1 - \frac{\hat{\mathbf{d}} \cdot \mathbf{d}^*}{\|\hat{\mathbf{d}}\|\|\mathbf{d}^*\|},
% \]
% where $\mathbf{d} = \mathbf{v} / \|\mathbf{v}\|$. To mitigate vanishing gradients when the predicted direction is opposite the ground truth, we additionally apply an auxiliary loss on the unnormalized direction logits. A temporal smoothness regularizer penalizes large angular deviations between consecutive predicted directions,
% \[
% \mathcal{L}_{\text{smooth}} = 1 - \cos(\hat{\mathbf{d}}_t, \mathbf{d}_{t-1}^*).
% \]
% The final training objective is a weighted sum of these components.
We supervise slip direction with a cosine similarity loss,
\[
\mathcal{L}_{\text{dir}} = 1 - \hat{\mathbf{d}}^\top \mathbf{d}^*,
\]
where $\hat{\mathbf{d}} = \hat{\mathbf{v}} / \|\hat{\mathbf{v}}\|$ and $\mathbf{d}^* = \mathbf{v}^* / \|\mathbf{v}^*\|$ are the predicted and ground-truth unit direction vectors, respectively. To mitigate vanishing gradients when the predicted direction opposes the ground truth, we additionally apply an auxiliary loss on the unnormalized direction logits $\hat{\mathbf{v}}$. We further include a temporal smoothness regularizer that penalizes large angular deviations between consecutive predicted directions,
\[
\mathcal{L}_{\text{smooth}} = 1 - \hat{\mathbf{d}}_t^\top \mathbf{d}_{t-1}^*,
\]
and define the final training objective as a weighted sum of all components, $
\mathcal{L} = \lambda_{\text{slip}}\,\mathcal{L}_{\text{slip}} + \lambda_{\text{mag}}\,\mathcal{L}_{\text{mag}} + \lambda_{\text{dir}}\,\mathcal{L}_{\text{dir}} + \lambda_{\text{smooth}}\,\mathcal{L}_{\text{smooth}}.$
In our experiments, we set $\lambda_{\text{dir}} = 2.0$ to prioritize direction estimation, $\lambda_{\text{slip}} = 1.0$, $\lambda_{\text{mag}} = 0.5$, and $\lambda_{\text{smooth}} = 0.1$, treating the smoothness term as a light regularizer.

% \subsection{Training Procedure}
% We adopt a two-stage training strategy. First, the model is pretrained on an audio dataset of the robot self-slipping from a stationary calibration object to learn a general acoustic representation of slip. Second, the pretrained model is finetuned on task-specific slip data with precise motion capture supervision to adapt the model to the target sensing setup with object slip induced by the environment. During finetuning, the pretrained audio encoder is frozen and only the task-specific prediction heads are optimized. This preserves the learned acoustic representation while enabling specialization of slip detection, magnitude estimation, and direction prediction for the target domain.

% The model is trained using the Adam optimizer with a learning rate of $10^{-3}$ and weight decay of $10^{-4}$. To improve robustness across surface textures and contact conditions, we apply SpecAugment-style time and frequency masking~\cite{park2019specaugment} as well as random gain augmentation. Frames without slip are subsampled or reweighted to mitigate class imbalance. Models are trained for up to 1000 epochs with early stopping based on validation loss. At inference time, predictions with $p(\text{slip}) < 0.5$ are mapped to a zero slip vector.

\subsection{Data Collection and Model Training}

Learning slip direction and magnitude from audio can require both large amounts of labeled data and accurate ground-truth supervision. Since collecting large-scale datasets with precise slip labels during real manipulation needs specialized external tracking systems, we adopt a two-stage data collection and training strategy that separates representation learning from task-specific adaptation.

We mount the parallel-jaw gripper equipped with the \algname sensor on a robot arm and use a motion capture system calibrated to the gripper's grasp plane to precisely track the in-plane motion of both the gripper and the object, allowing slip to be inferred from their relative poses (Fig.~\ref{fig:system}). First, we collect an audio dataset of \textit{robot-induced slip} by executing randomized robot motions that sweep the gripper across a stationary calibrated 3D printed probe (Fig.~\ref{fig:data_collection}, ``Robot-Induced Slip''). For this dataset, we compute slip direction and magnitude labels directly from the recorded robot state. Second, we collect a small dataset of \textit{externally-induced slip} by manually perturbing five objects (four from the YCB dataset~\cite{calli2015ycb}) grasped by a stationary gripper (Fig.~\ref{fig:data_collection}, ``Externally-Induced Slip''). For each object, we record 30-second trials while using a motion capture system to automatically obtain slip direction and magnitude labels. We perform 30 trials with the robot on and stationary, 10 with the robot executing random motions, and 20 with the robot off.

We use the \textit{robot-induced slip} dataset to pretrain the model to learn a general acoustic representation of slip. We then finetune the pretrained model on the \textit{externally-induced slip} dataset to adapt to the target sensing scenario, where slip arises from external disturbances. During finetuning, we freeze the audio encoder and optimize only the task-specific prediction heads. This preserves the learned acoustic representation while enabling specialization for slip detection, magnitude estimation, and direction prediction.

We train the model using the Adam optimizer with a learning rate of $10^{-3}$ and weight decay of $10^{-4}$. To improve robustness across surface textures and contact conditions, we apply SpecAugment-style time and frequency masking~\cite{park2019specaugment} as well as random gain augmentation. To mitigate class imbalance, we subsample and reweight frames without slip. We train the models for up to 1000 epochs, stopping early based on validation loss. Inference treats predictions with $p(\text{slip}) < 0.5$ as a zero slip vector.

\input{floats/fig_data}

\section{Evaluation}
We evaluate \algname{} against baselines and system ablations to isolate contributions of the training regime, microphone number and placement, and spectrogram temporal window size. Additionally, we evaluate robustness to robot operating noise, and system integration into closed-loop reactive control tasks. All experiments use a parallel-jaw gripper with \algname{} sensors and objects tracked by an external motion capture system to provide the ground truth. 

\input{floats/tab_ablation}

\subsection{Baselines}{\label{sec:system_ablation}}
We compare against baselines from prior work and against several system ablations. Prior studies on embedding low-profile microphones into end-effectors commonly use SVM regressors~\cite{zoller2020active} or single-microphone sensing for event detection~\cite{zhang2025misalignment}. In Table~\ref{tab:controlled_slip}, we compare the \algname{} model against an SVM baseline and against single-microphone variants trained with and without pretraining. \algname{} model achieves the best overall performance, improving detection accuracy by up to $12\%$ and reducing directional MAE by up to $32\%$ relative to these baselines. Compared specifically to the single-microphone variants, the 4-microphone finetuned model reduces directional error by $64\%$ and magnitude RMSE by $68\%$, showing that the gains come from both combining \algname's training procedures with spatially distributed multi-channel sensing.

Additionally, we compare four microphone configurations in Fig.~\ref{fig:design}. Among the 2-microphone variants, microphone placement influences performance by affecting how well the sensor captures vibration asymmetries across the grasp. \algname{} (2-mic, corners, finetuned) achieves the best performance among the 2-microphone configurations, reducing directional MAE by approximately 5\% relative to \algname{} (2-mic, same finger, finetuned) and by about 13\% relative to \algname{} (2-mic, centered, finetuned). This suggests that distributing microphones across opposite fingers helps preserve bilateral differences in vibration propagation that encode slip direction. In contrast, when both microphones are placed on a single finger, the model cannot directly observe cross-finger vibration differences, which weakens the directional signal available for inference. Even with bilateral sensing, the performance gap between the corners and centered placements indicates that increasing the spatial baseline between microphones further improves sensitivity to directional vibration patterns. Expanding to \algname{} (4-mic, finetuned) provides an additional 22\% reduction in directional MAE relative to the best 2-microphone configuration and improves slip detection accuracy by 14\%. These gains suggest that denser spatial sampling of the vibration field allows the model to learn more stable cross-channel relationships associated with slip direction, while magnitude estimation appears largely governed by overall vibration energy and therefore benefits less from additional channels.

The bottom rows of Table~\ref{tab:controlled_slip} evaluate the effect of temporal window size by comparing spectrogram windows of 100\,ms, 200\,ms, and 300\,ms. Increasing the window size consistently improves slip detection accuracy, indicating that longer temporal context makes it easier for the model to distinguish sustained slip events from transient contact noise. At the same time, both directional and magnitude estimation degrade as the window grows longer. A likely explanation is that slip direction and intensity often vary within a single window, particularly during externally induced disturbances, and aggregating over longer time intervals blurs these instantaneous dynamics. Shorter windows better preserve the local structure of the slip signal but provide less evidence for reliably detecting whether slip is occurring. The 200\,ms window represents a balance between these effects. %: it provides sufficient temporal context to improve slip detection while still preserving enough temporal resolution to estimate slip direction and magnitude with reasonable accuracy.
% \textbf{Model Finetuning Ablation.}
% We compare three training regimes: no pretraining, pretraining only, and the full \algname pipeline with pretrained encoder frozen and prediction heads finetuned with motion capture labels.

\input{floats/tab_objects}

\subsection{Cross-Object Generalization}
Table~\ref{tab:per_object} reports per-object results using the 4-microphone configuration to assess sensitivity to object geometry and surface material. We compare models trained on each object individually (per-obj.) against a single model trained on all objects jointly (all-obj.). The joint model achieves comparable or improved directional accuracy across all objects, reducing directional MAE by approximately 2\%--9\% relative to the per-object specialist models for Glass Cleaner, Chips Can, Mustard Container, and Cracker Box, while maintaining nearly identical detection accuracy. These results suggest that training across diverse contact surfaces improves the robustness of the learned acoustic representation without sacrificing object-specific performance. Magnitude RMSE remains largely unchanged across objects and training regimes, indicating that slip magnitude estimation is relatively invariant compared to directional estimation.

\subsection{Slip with Robot Noise}
A concern for acoustic sensing is whether vibrations generated by the robot itself interfere with the slip signal. To isolate this effect, we evaluate each \emph{finetuned} model on the robot-induced pretraining validation set, where the robot actively executes the slip motion and the audio contains both slip-induced vibrations and robot operating noise. Robot noise does not substantially degrade performance for the 4-microphone model. It achieves a directional MAE of $15.9 \pm 16.6$ degrees and a magnitude RMSE of $0.5 \pm 0.2$ mm. Compared with its externally induced performance, the directional error increases by only $12.8\%$, suggesting that robot operating sound is not the dominant source of error for the best-performing model. The relative ordering across microphone layouts is also consistent. Among the 2-microphone variants, the centered layout is most sensitive to robot noise, reaching $39.2 \pm 26.0$ degrees directional MAE and $0.7 \pm 0.4$ mm magnitude RMSE, corresponding to a $92.2\%$ increase in directional error relative to the externally induced slip setting. In contrast, the corners layout reaches $21.3 \pm 20.9$ degrees and $0.5 \pm 0.4$ mm, only a $17.7\%$ increase in directional error, while the same-finger layout achieves $17.4 \pm 18.7$ degrees and $0.6 \pm 0.3$ mm, an $8.4\%$ reduction in directional MAE. These results suggest that configurations with spatial coverage remain accurate under active robot motion, whereas the centered 2-microphone layout degrades. %This complements the externally induced slip setting, in which the robot remains stationary and the disturbance comes from outside the grasp.

\input{floats/fig_evaluation}

\subsection{Reactive Control}

\input{floats/tab_task1}

We evaluate \algname in two closed-loop control tasks, where real-time audio is streamed to the model and predictions directly drive robot motion. 

In the first task (Fig.~\ref{fig:experiment}, left), the robot pushes a grasped object against a wall and stops upon detecting in-hand slip. We compare the \algname model with the SVM baseline over 10 trials per object. In each trial, the robot begins with the object in contact with the wall, retracts 10\,cm, and re-approaches. We measure the stopping error along the motion direction between the robot’s initial wall-contact pose and final stopping pose, denoted as $\Delta x$, and we consider a trial successful only if the object makes contact with the wall and the robot stops before pushing it past its full length (Table~\ref{tab:slip_stop}, left). Overall, \algname achieves a 100\,\% success rate, while the SVM baseline succeeds in 62\,\% of the trials and yields a mean stopping error 81.4\,\% larger than \algname. The SVM baseline also shows higher variance, largely driven by failure cases, and varies significantly across objects, indicating limited robustness to different contact conditions. These results suggest that \algname provides more reliable slip detection in closed-loop control, reducing both missed slip events and unstable behavior caused by incorrect predictions.

In the second task (Fig.~\ref{fig:experiment}, right), as an experimenter induces slip, the robot continuously tracks the predicted slip vector to maintain a stable object-gripper relative pose. We perform 10 trials for each object. Qualitatively, \algname enables the robot to reliably follow the object as it moves, whereas the SVM baseline often fails to produce meaningful motion due to inaccurate slip predictions. Quantitatively, we evaluate how well the gripper maintains a constant relative pose with the object in the grasp plane by computing the RMSE of gripper-object poses along the trajectory (Table~\ref{tab:slip_stop}, right). Overall, \algname achieves an RMSE that is 50.5\,\% lower than the SVM baseline. These results indicate that \algname{} predicts slip direction and magnitude with sufficient accuracy to enable fast and reliable robot reactions to in-hand slip, supporting real-time feedback-based tracking.
%The baseline error remains moderate mainly because the object can move only within a limited region while still maintaining contact with the gripper. 

% Let $\mathbf{p}_g(t)$ and $\mathbf{p}_o(t)$ denote the gripper and object positions at time $t$, respectively. We define a reference offset $\mathbf{d}_{ref}$ as the median relative position during the initial 0.1\,s of each trial, and compute the error
%$\mathbf{e}(t) = (\mathbf{p}_g(t) - \mathbf{p}_o(t)) - \mathbf{d}_{ref}$.

%Qualitatively, the SVM baseline rarely produces meaningful robot motion due to inaccurate slip direction and magnitude prediction, whereas \

% \begin{figure}[t!]
%     \centering
%     \includegraphics[width=\linewidth]{figures/evaluation.pdf}
% \caption{\textbf{Qualitative Evaluation of \algname.} Each row shows a different object; each column shows a sample evaluation frame with predicted (gray) and ground-truth (green) slip vectors overlaid on the contact image alongside per-channel log-mel spectrograms. \algname accurately estimates slip direction and magnitude across objects with varying geometry and surface material, even under impulsive externally induced slip.}
%     \label{fig:evaluation}
% \end{figure}

\section{Limitations and Future Work}
\algname has several limitations that suggest directions for future work. The system estimates planar slip only and does not model rotational slip about the grasp axis; extending the slip representation to include rotational components would provide more complete coverage of in-hand motion. Although the textured silicone pad promotes structured vibrations, acoustic signatures vary with object surface material and fully smooth or compliant objects may produce weaker signals and degrade accuracy, motivating domain adaptation or online recalibration strategies. The finetuning stage relies on motion capture for ground-truth labels, which may be unavailable in many settings; self-supervised or weakly supervised labeling would reduce this dependency. Evaluation is limited to a single parallel-jaw gripper, and differences in finger geometry or material may require adaptation. The 200\,ms inference window introduces latency that could limit performance in high-speed tasks, suggesting further exploration of shorter windows with history of observations or causal streaming architectures.

\input{floats/fig_task}

\section{Conclusion}
We present \algname, an acoustic sensing system for continuous planar slip vector estimation in robotic in-hand manipulation. By embedding low-cost piezoelectric microphones behind textured silicone contact pads on a parallel-jaw gripper, \algname captures structure-borne vibrations induced by gripper-object slip without requiring cameras, optics, or complex fabrication. Our slip prediction network processes synchronized multi-channel log-mel spectrograms using a convolutional architecture with learned channel attention and temporal attention pooling, jointly estimating slip presence, magnitude, and direction within a unified multi-objective framework. A two-stage training strategy that combines pretraining on robot-induced slip data with finetuning enables the model to learn transferable acoustic slip representations and adapt them to task-specific conditions.

Experimental results show that \algname achieves strong performance in slip detection, direction estimation, and magnitude regression. In particular, the finetuned 4-microphone configuration outperforms all baselines, including the SVM baseline and pretraining-only variants, across all evaluation metrics. These results demonstrate that multi-channel acoustic sensing, when combined with learning-based fusion, provides a practical and effective solution for continuous slip vector estimation required for closed-loop grasp correction. Overall, \algname strengthens acoustic sensing as a compelling modality for slip estimation, offering advantages in form factor, durability, and deployment cost.

\section*{Acknowledgments}
This work was supported by Samsung Research America and NSF Graduate Research Fellowship under Grant No. DGE2140739.

\bibliographystyle{ieeetr}
\bibliography{references.bib}

\end{document}

%% file: floats/fig_front.tex
\begin{figure}[t!]
    \centering
    \includegraphics[width=\linewidth]{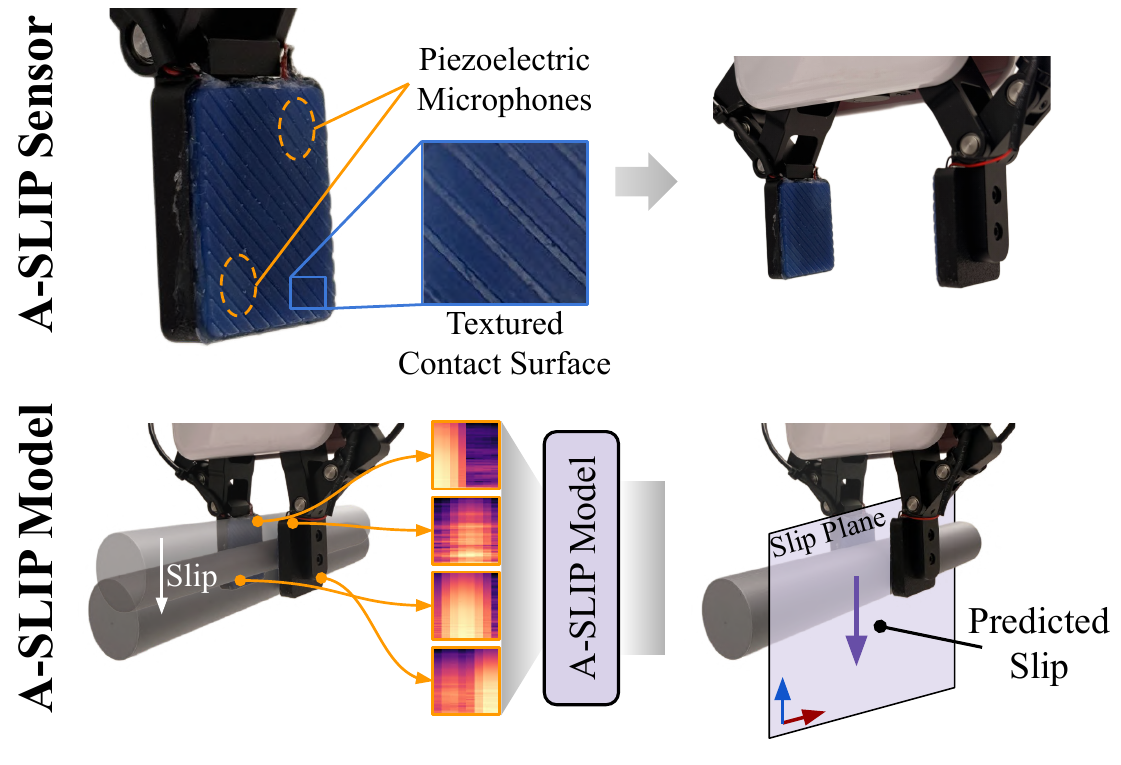}
    \caption{\textbf{Overview of A-SLIP:} Piezoelectric microphones embedded behind textured silicone contact pads capture structure-borne vibrations during slip. Multi-channel log-mel spectrograms are processed by a convolutional network with channel and temporal attention to jointly estimate slip presence, magnitude, and direction as $\mathbf{v}_t \in \mathbb{R}^2$.
    }
    \label{fig:front}
    \vspace{-0.3cm}
\end{figure}

%% file: floats/fig_design.tex
\begin{figure}[t!]
    \centering
    \includegraphics[width=\linewidth]{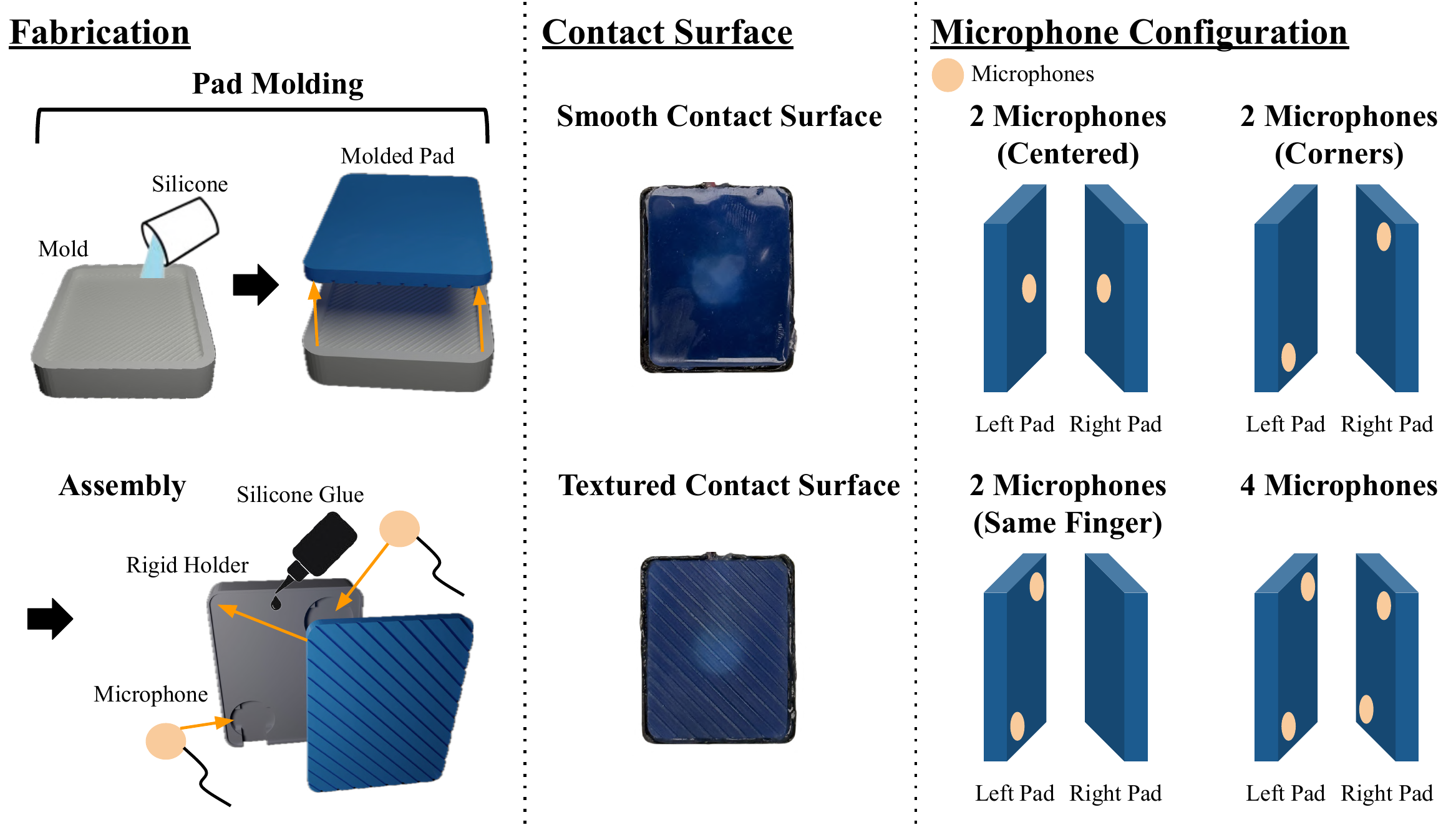}
    \caption{%
    \textbf{Design of the \algname{} Sensor.}
    (\textit{Left}) Fabrication pipeline: silicone is cast in a 3D-printed mold to form a compliant contact pad, which is bonded to a rigid holder with embedded piezoelectric microphones.
    (\textit{Center}) Contact surface variants: smooth and textured silicone pads; the textured surface introduces controlled asperities that produce more directionally informative vibrations during slip.
    (\textit{Right}) Evaluated microphone placements: three two-microphone layouts (centered, corners, and same-finger) and a four-microphone layout with two microphones per finger to increase contact-region coverage.
    }
    % \caption{%
    % \textbf{Design of the \algname{} sensor.} (\textit{Top}) Silicone rubber is cast in a 3D-printed mold to form a contact pad, bonded to a rigid holder with embedded piezoelectric microphones. (\textit{Bottom left}) Smooth and textured contact surfaces; texturing reduces directional ambiguity. (\textit{Bottom right}) Microphone placements: the \textit{2-microphone configuration} places one microphone per finger or both on the same finger; the \textit{4-microphone configuration} places two per finger for greater contact patch coverage.
    % }
    \label{fig:design}
    \vspace{-0.3cm}
\end{figure}

%% file: floats/fig_model.tex
\begin{figure}[t!]
    \centering
    \includegraphics[width=\linewidth]{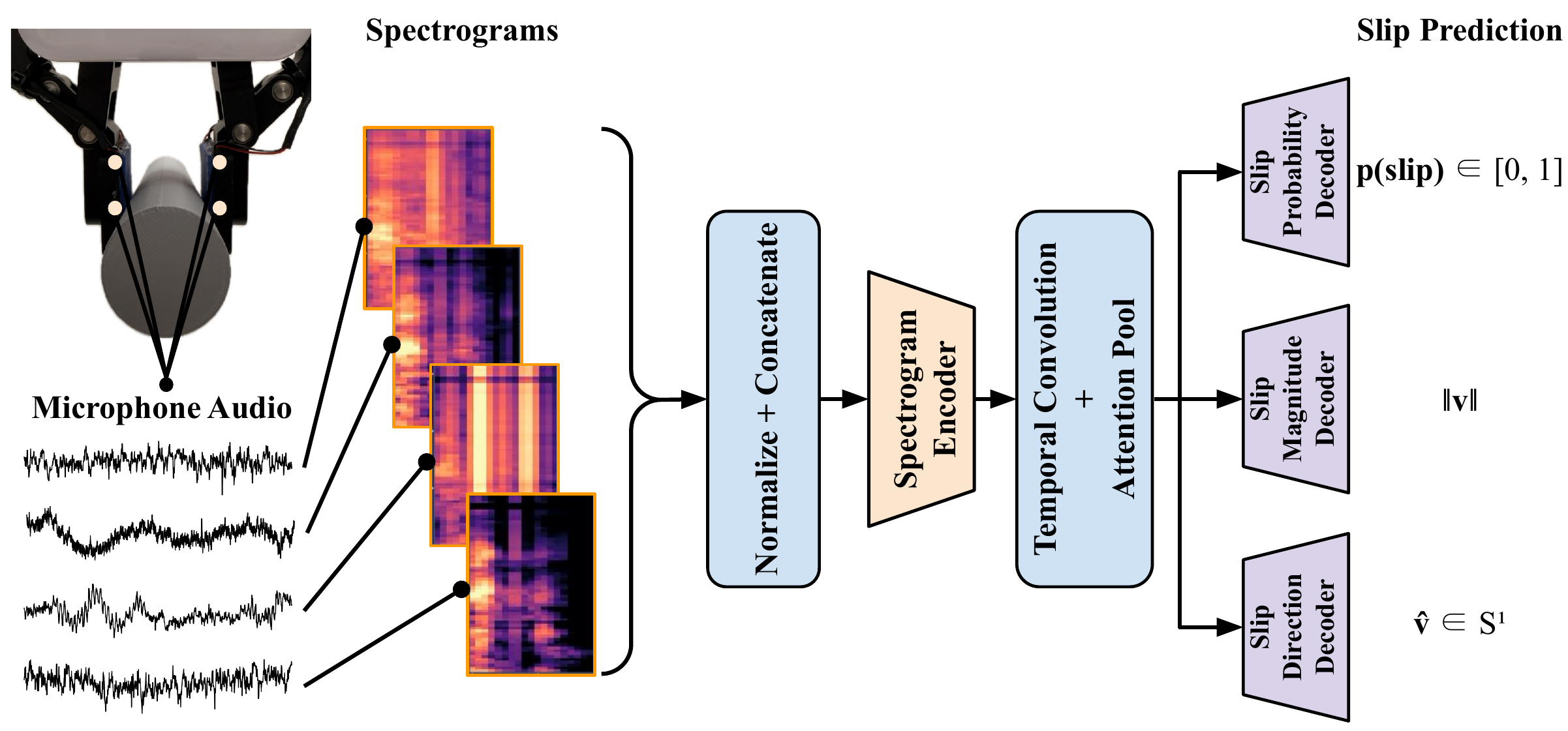}
    \caption{%
     \textbf{\algname{} Model Architecture.} Log-mel spectrograms from synchronized microphones are normalized and fused via a learned channel attention module. The fused representation passes through 2D convolutional layers preserving temporal resolution, then 1D temporal convolutions modeling slip dynamics. A temporal attention pooling module aggregates features into a latent vector passed to three heads: a \textit{slip classification head} predicting $p(\text{slip}) \in [0,1]$, a \textit{magnitude regression head} predicting $\|\mathbf{v}\|$, and a \textit{direction head} predicting $\hat{\mathbf{v}} \in \mathcal{S}^1$.
    }
\vspace{-0.3cm}
    \label{fig:architecture}
\end{figure}

%% file: floats/fig_system.tex
\begin{figure}[t!]
    \centering
    \includegraphics[width=0.7\linewidth]{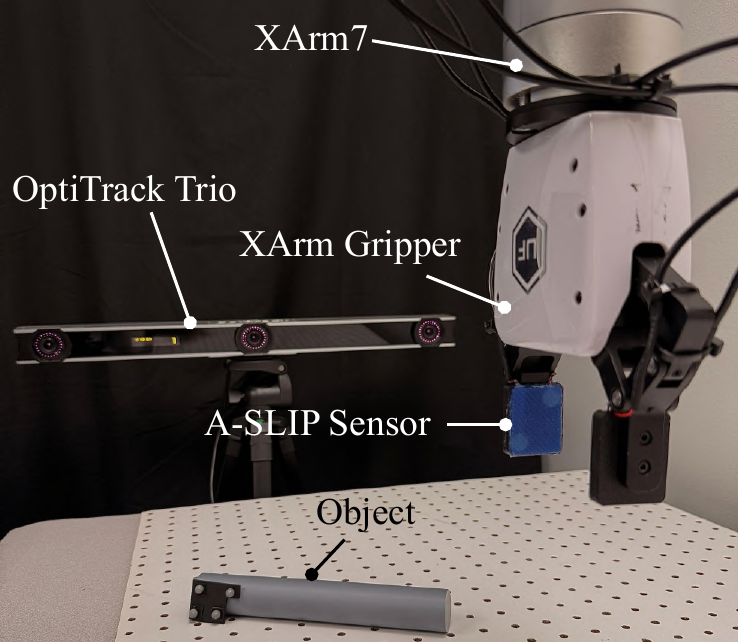}
    \caption{%
    %\vspace{-0.2cm}
    \textbf{\algname System.} We mount A-SLIP sensors on an XArm gripper attached to an XArm7 robot. To obtain ground-truth in-hand slip, we use an OptiTrack Trio to track poses of the left finger and the object, each with reflective markers attached.
    }
    \label{fig:system}
    \vspace{-0.3cm}
\end{figure}

%% file: floats/fig_data.tex
% \begin{figure}[t!]
%     \centering
%     \includegraphics[width=0.9\linewidth]{figures/training_data_v3.pdf}
%     \caption{%
%     %\vspace{-0.2cm}
%     \textbf{\algname Training Data Distribution} (\textit{Top}) We collect the pretraining dataset automatically by executing randomized robot motions that sweep the gripper across a stationary 3D-printed probe and derive slip direction and magnitude from robot states. (\textit{Bottom}) We collect the finetuning dataset by manually perturbing objects grasped by a stationary gripper and compute slip labels from the relative rigid-body motion tracked by OptiTrack. Counts indicate the number of 200\,ms data slices in each dataset.
%     }
%     \label{fig:data_collection}
%     \vspace{-0.6cm}
% \end{figure}

\begin{figure}[t!]
    \centering
    \includegraphics[width=0.9\linewidth]{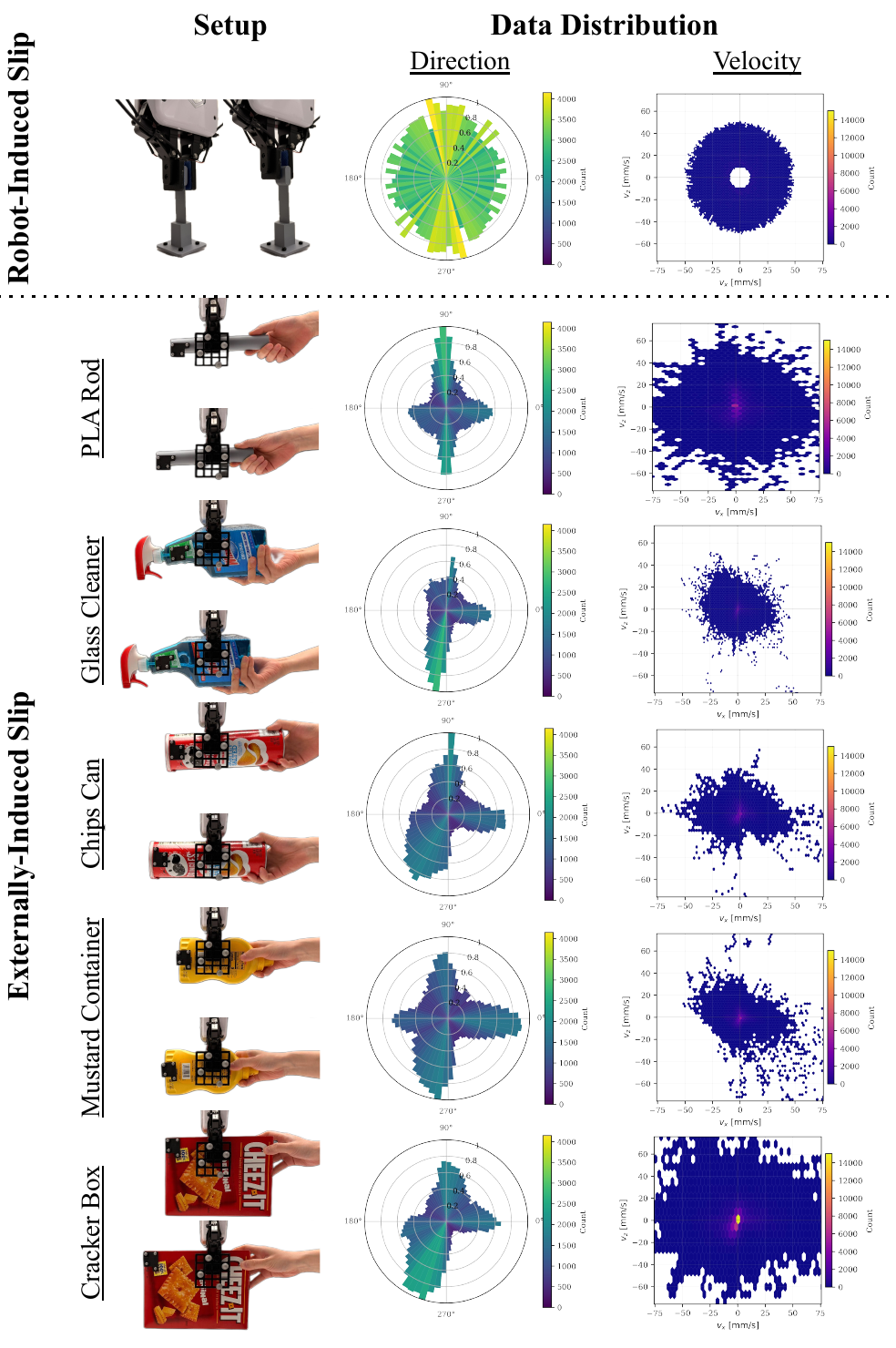}
    \caption{%
    \textbf{\algname Dataset Distribution.} (\textit{Top}) Robot-induced slip data collected automatically via randomized robot motions sweeping the gripper across a stationary probe, with labels derived from robot states. (\textit{Bottom}) Externally-induced slip data collected by manually perturbing grasped objects, with labels from OptiTrack-tracked rigid-body motion. Counts indicate the number of audio slices in the dataset.
    }
    \label{fig:data_collection}
    \vspace{-0.3cm}
\end{figure}

%% file: floats/tab_ablation.tex
\begin{table}[t]
  \centering
  \caption{Slip prediction model comparison. Binary slip accuracy (Det.). Dir.\ MAE: Slip direction MAE (Dir.\ MAE). Slip magnitude RMSE (Mag.\ RMSE).}
  \label{tab:controlled_slip}
  % \renewcommand{\arraystretch}{1.15}
  %\resizebox{\columnwidth}{!}{%
  \scriptsize
  \begin{tabular}{@{}l@{\quad}ccc@{}}
    \toprule
    \multirow{2}{*}{\textbf{Method}} & \textbf{Det.} & \textbf{Dir.\ MAE} & \textbf{Mag.\ RMSE} \\
    & (\%) & (deg) & (mm) \\
    \midrule
  SVM& 73.0 & $19.2 \pm 28.6$ & $1.9 \pm 2.4$ \\
    \midrule
  Single mic (no pretrain) & 43.3 & $28.5 \pm 26.5$ & $2.0 \pm 1.0$ \\
  Single mic (pretrain + finetune)  & 70.2 & $20.7 \pm 22.4$ & $2.7 \pm 0.9$ \\
    \midrule
  Pretrain only (2-mic, centered)  & 60.4 & $29.3 \pm 21.5$ & $1.5 \pm 1.2$ \\
  Pretrain only (2-mic, corners)& 63.1 & $26.8 \pm 21.1$ & $1.5 \pm 1.2$ \\
  Pretrain only (2-mic, same finger)& 61.2 & $26.7 \pm 21.2$ & $1.6 \pm 1.2$ \\
  Pretrain only (4-mic)& 72.4 & $21.6 \pm 8.0$ & $3.2 \pm 1.4$ \\
    \midrule
  \algname (2-mic, centered, finetuned)& 63.6 & $20.4 \pm 16.0$ & $1.0 \pm 1.3$ \\
  \algname (2-mic, corners, finetuned)& 71.9 & $18.1 \pm 9.3$ & $0.6 \pm 0.6$ \\
  \algname (2-mic, same finger, finetuned)& 72.1 & $19.0 \pm 10.1$ & $0.7 \pm 0.5$ \\
  \textbf{\algname (4-mic, finetuned)}& \textbf{81.8} & $\mathbf{14.1 \pm 6.9}$ & $1.0 \pm 0.9$ \\
    \midrule
  \algname\ (100\,ms window)& 72.5 & $\mathbf{8.2 \pm 9.6}$ & $\mathbf{0.6 \pm 0.7}$ \\
  \algname\ (200\,ms window)& 81.8 & $14.1 \pm 6.9$ & $1.0 \pm 0.9$ \\
  \algname\ (300\,ms window)& \textbf{88.5} & $20.9 \pm 6.5$ & $1.4 \pm 1.1$ \\
    \bottomrule
  \end{tabular}
\end{table}

%% file: floats/tab_objects.tex
\begin{table}[t]
\label{tab:per_object}
  \centering
  \caption{Per-object slip prediction results. Binary slip accuracy (Det.). Slip direction MAE (Dir.\ MAE). Slip magnitude RMSE (Mag.\ RMSE).}
  \label{tab:per_object}
  % \renewcommand{\arraystretch}{1.15}
  % \resizebox{\columnwidth}{!}{%
  \scriptsize
  \begin{tabular}{@{}lcccc@{}}
    \toprule
    \multirow{2}{*}{\textbf{Object}} &
    \multirow{2}{*}{\textbf{Train}} &
    \textbf{Det.} & \textbf{Dir. MAE} & \textbf{Mag. RMSE} \\
    & & (\%) & (deg) & (mm) \\
    \midrule
    \multirow{2}{*}{PLA Rod}
        & Per-obj. & 94.5 & $26.07 \pm 9.8$ & $\mathbf{2.44 \pm 1.20}$ \\
        & All-obj. & \textbf{94.7} & $\mathbf{25.58 \pm 9.1}$ & $2.72 \pm 1.30$ \\
    \midrule
    \multirow{2}{*}{Glass Cleaner}
        & Per-obj. & \textbf{81.1} & $15.17 \pm 7.4$ & $0.60 \pm 0.72$ \\
        & All-obj. & 78.7 & $\mathbf{14.38 \pm 6.8}$ & $\mathbf{0.58 \pm 0.68}$ \\
    \midrule
    \multirow{2}{*}{Chips Can}
        & Per-obj. & \textbf{71.8} & $9.01 \pm 6.1$ & $\mathbf{0.40 \pm 0.63}$ \\
        & All-obj. & 71.0 & $\mathbf{8.58 \pm 6.5}$ & $0.41 \pm 0.66$ \\
    \midrule
    \multirow{2}{*}{Mustard Container}
        & Per-obj. & \textbf{84.0} & $24.14 \pm 9.6$ & $0.62 \pm 0.82$ \\
        & All-obj. & 83.7 & $\mathbf{22.20 \pm 8.9}$ & $\mathbf{0.60 \pm 0.80}$ \\
    \midrule
    \multirow{2}{*}{Cracker Box}
        & Per-obj. & 79.4 & $7.61 \pm 5.9$ & $0.62 \pm 0.74$ \\
        & All-obj. & \textbf{80.6} & $\mathbf{6.90 \pm 5.5}$ & $\mathbf{0.55 \pm 0.70}$ \\
    \bottomrule
  \end{tabular}
  
\end{table}

%% file: floats/fig_evaluation.tex
\begin{figure}[t!]
    \centering
    \includegraphics[width=\linewidth]{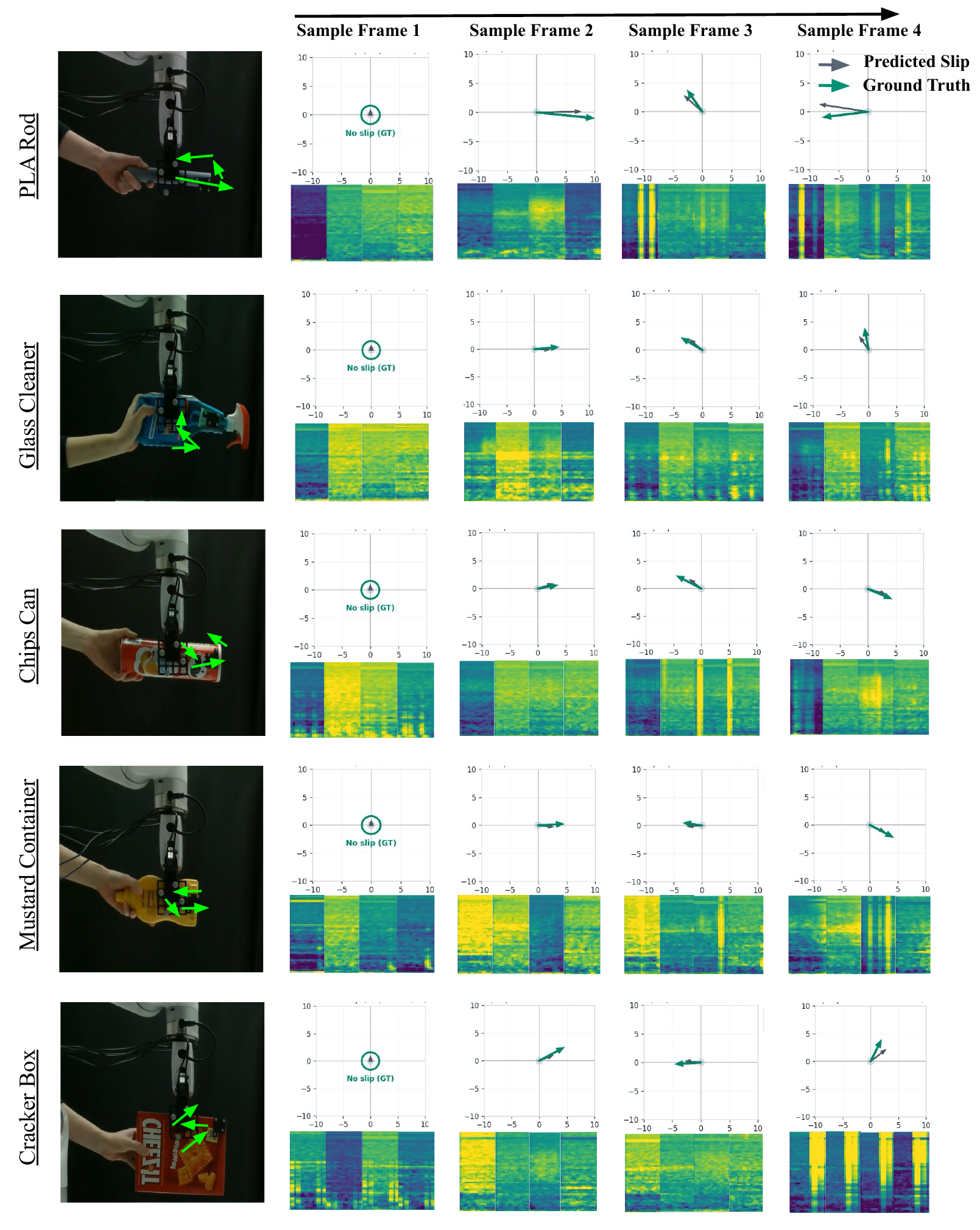}
\caption{\textbf{Qualitative Evaluation of \algname.} Each row shows a different object; each column shows a sample evaluation frame with predicted (gray) and ground-truth (green) slip vectors overlaid on the contact image alongside per-channel log-mel spectrograms. \algname accurately estimates slip direction and magnitude across objects with varying geometry and surface material, even under impulsive externally induced slip.}
    \label{fig:evaluation}
    \vspace{-0.3cm}
\end{figure}

%% file: floats/tab_task1.tex
\begin{table}[t]
\centering
\caption{Task performances. Success rate (Succ.), stopping error ($\Delta x$), and RMSE of gripper-object pose along trajectory (Pose RMSE).}
\label{tab:slip_stop}
\scriptsize
\setlength{\tabcolsep}{3pt}

\resizebox{\columnwidth}{!}{
\begin{tabular}{@{}lcccc@{}}
\toprule
 & \multicolumn{2}{c}{\textbf{Task 1: Slip-Stop}} & \multicolumn{2}{c}{\textbf{Task 2: Slip-Track}} \\
\cmidrule(lr){2-3} \cmidrule(lr){4-5}
 & \multicolumn{2}{c}{\textbf{Succ.}, $\mathbf{\Delta x}$ (mm)} & \multicolumn{2}{c}{\textbf{Pose RMSE} (mm)} \\
\cmidrule(lr){2-3} \cmidrule(lr){4-5}
\textbf{Object} & SVM & \algname & SVM & \algname \\
\midrule

PLA Rod & 7/10, $39.7 \pm 38.7$ & \textbf{10/10}, $\mathbf{16.7 \pm 8.9}$ & $32.4 \pm 20.2$ & $\mathbf{16.9 \pm 12.4}$ \\

Glass Cleaner & 6/10, $13.4 \pm 18.7$ & \textbf{10/10}, $\mathbf{10.5 \pm 5.7}$ & $32.4 \pm 18.1$ & $\mathbf{15.0 \pm 9.5}$ \\

Chips Can & 8/10, $\mathbf{16.7 \pm 16.4}$ & \textbf{10/10}, $19.8 \pm 6.6$ & $33.8 \pm 21.7$ & $\mathbf{12.9 \pm 9.4}$ \\

Mustard Container & 5/10, $17.4 \pm 31.3$ & \textbf{10/10}, $\mathbf{9.7 \pm 6.7}$ & $25.0 \pm 14.4$ & $\mathbf{20.5 \pm 14.4}$ \\

Cracker Box & 5/10, $44.3 \pm 27.6$ & \textbf{10/10}, $\mathbf{15.8 \pm 7.1}$ & $35.5 \pm 20.4$ & $\mathbf{13.0 \pm 9.3}$ \\

\midrule

\textbf{Overall} & 31/50, $26.3 \pm 29.6$ & \textbf{50/50}, $\mathbf{14.5 \pm 7.8}$ & $32.0 \pm 19.3$ & $\mathbf{15.8 \pm 11.3}$ \\

\bottomrule
\end{tabular}
}
\end{table}

%% file: floats/fig_task.tex
\begin{figure}[t!]
    \centering
    \includegraphics[width=\linewidth]{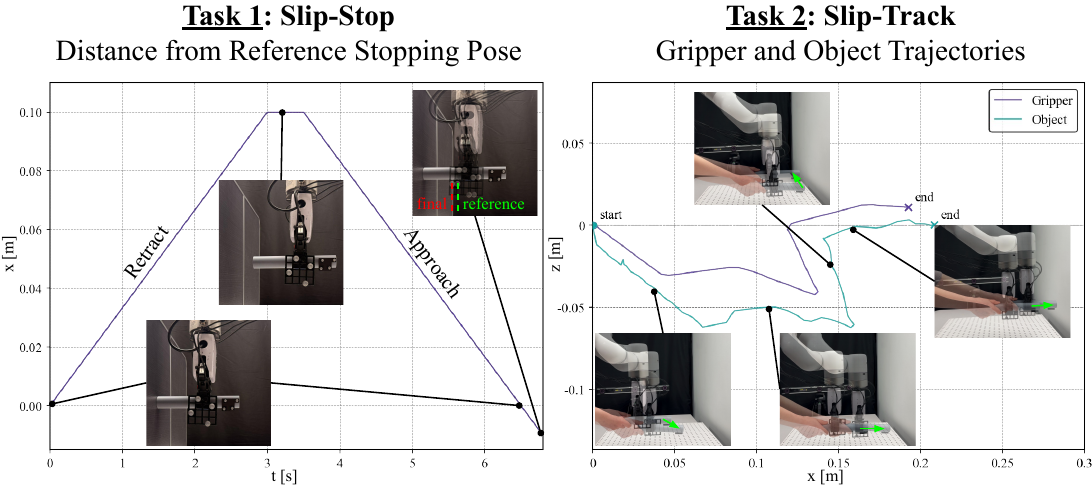}
    \caption{%
    %\vspace{-0.2cm}
    \textbf{Reactive Control.} \algname predicts slip direction and magnitude in real time, enabling rapid robot responses to in-hand slip. (\textit{Left}) Task 1: the robot pushes an object against a wall and stops automatically upon in-hand slip detection. (\textit{Right}) Task 2: as an experimenter induces slip, the robot follows the model-predicted slip vector to maintain a stable grasp.
    }
    \label{fig:experiment}
    \vspace{-0.3cm}
\end{figure}